\documentclass[11pt]{article}

\usepackage{amsmath,amssymb}
\usepackage{changepage}
\usepackage[utf8x]{inputenc}
\usepackage{cite}
\usepackage{nameref,hyperref}
\usepackage{microtype}
\usepackage[table]{xcolor}
\usepackage{array}
\usepackage{authblk}
\usepackage{lastpage,fancyhdr,graphicx}
\usepackage[margin=0.70in]{geometry}

\newcommand{\Probab}[1]{\mathcal{P}({#1})}
\newcommand{\Pcond}[2]{\Probab{{#1} \mid{#2}}}

\newcommand{\bth}{\boldsymbol{\theta}}

\usepackage[aboveskip=1pt,labelfont=bf,labelsep=period,justification=raggedright,singlelinecheck=off]{caption}

\begin{document}

\title{Investigating the performance of multi-objective optimization when learning Bayesian Networks}

\author[1,2,3]{Marco S. Nobile}
\author[2,4]{Paolo Cazzaniga}
\author[5,6]{Daniele Ramazzotti\footnote{Corresponding author: daniele.ramazzotti@unimib.it}}
\affil[1]{Department of Informatics, Systems and Communication, University of Milano-Bicocca, Milan, Italy}
\affil[2]{SYSBIO/ISBE.IT Centre for Systems Biology, Milan, Italy}
\affil[3]{Department of Industrial Engineering \& Innovation Sciences, Eindhoven University of Technology, Eindhoven, The Netherlands}
\affil[4]{Department of Human and Social Sciences, University of Bergamo, Bergamo, Italy}
\affil[5]{Department of Pathology, Stanford University, CA, United States}
\affil[6]{Department of Medicine and Surgery, University of Milano-Bicocca, Milano, Italy} 
\date{}

\maketitle 

\section*{Abstract}
Bayesian Networks have been widely used in the last decades in many fields, to describe statistical dependencies among random variables. 
In general, learning the structure of such models is a problem with considerable theoretical interest that poses many challenges. 
On the one hand, it is a well-known NP-complete problem, practically hardened by the huge search space of possible solutions. 
On the other hand, the phenomenon of I-equivalence, i.e., different graphical structures underpinning the same set of statistical dependencies, may lead to multimodal fitness landscapes further hindering maximum likelihood approaches to solve the task. 
Despite all these difficulties, greedy search methods based on a likelihood score coupled with a \textcolor{black}{regularizator score} to account for model complexity, have been shown to be surprisingly effective in practice.
In this paper, we consider the formulation of the task of learning the structure of Bayesian Networks as an optimization problem based on a likelihood score, without complexity terms to regularize it.
In particular, we exploit the NSGA-II multi-objective optimization procedure in order to explicitly account for both the likelihood of a solution and the number of selected arcs, by setting these as the two objective functions of the method.
The aim of this work is to investigate the behavior of NSGA-II and analyse the quality of its solutions.
We thus thoroughly examined the optimization results obtained on a wide set of simulated data, by considering both the goodness of the inferred solutions in terms of the objective functions values achieved, and by comparing the retrieved structures with the ground truth, i.e., the networks used to generate the target data. 
Our results show that NSGA-II can converge to solutions characterized by better likelihood and less arcs than classic approaches, although paradoxically characterized in many cases by a lower similarity with the target network.

\section*{Introduction}
 \label{sec:introduction}
Bayesian Networks (BNs)\cite{koller2009probabilistic} are widely used to provide a succinct description of the statistical dependencies among random variables. 
Their applications are manifold, ranging from diagnostics, discovery of gene regulatory networks, genetic programming and many other tasks \cite{hasegawa2008bayesian,caravagna2016algorithmic,bonchi2017exposing,myte2017untangling,gendelman2017bayesian,cai2017analysis,mcnally2017co,zhang2017analysis,gao2018causal,agrahari2018applications}.
In any case, learning the structure of a BN is a non trivial task and still poses many challenges, both from a theoretical and practical standpoint. 
The state-of-the-art approaches to tackle this problem are mainly of two kinds\footnote{Recently, hybrid approaches, e.g., combining the two classic formulations, have been developed. This third kind of models can be very effective in domain specific contexts, but, for the sake of clarity, here we focus on the classic formulations of the problem of learning the structure of BNs\cite{koller2009probabilistic}.}. 
The \emph{constraint-based} techniques, where the BN is formalized as a set of relations of conditional dependency among random variables to be, in turn, learned. 
These methods typically provide a \emph{causal} interpretation of the underlying structure \cite{pearl2009causality}. 
The \emph{score-based} techniques do not attempt to give any causal interpretation to the network, but reformulate the task as an optimization problem with a fitness function usually based on likelihood adjusted with a complexity term \cite{koller2009probabilistic,gamez2011learning}. 
Regardless of the approach employed to learn a BN, this is a well-known NP-complete problem, due to the huge search space of possible solutions, which effectively prevents any exhaustive search for BNs with more than a few nodes \cite{robinson1977counting,chickering1996learning}. 

In addition, this problem is hardened by the phenomenon of I-equivalence, i.e., the occurrence of different graphical structures that can underpin the same set of statistical dependencies. 
I-equivalence may lead to multimodal fitness landscapes, further complicating maximum likelihood approaches. 
Because of this, any method for structure learning of BNs may converge to a set of equivalent optimal networks that, albeit structurally different, subsume the same induced distribution over the variables \cite{koller2009probabilistic}. 

Despite all the difficulties mentioned above, greedy search methods based on a likelihood score coupled with a \textcolor{black}{regularizator} term to account for model complexity, have been shown to be surprisingly effective in practice \cite{gamez2011learning,beretta2017quantitative,ramazzotti2017learning,scutari2019learns}.
In this work, we consider a specific methodology that can be placed within the score-based approaches, in which the task of learning the structure of a BN is formalized as an optimization problem. 
In particular, we do not make use of a score composed by both a likelihood term and a \textcolor{black}{regularizator} to penalize complexity; instead, we directly account for the complexity of the solutions by means of a multi-objective optimization technique. 
Such formulation \textcolor{black}{allows us to explicitly take into account the} trade-off between likelihood and model complexity. 
Specifically, we adopt a multi-objective optimization algorithm, called Non-dominated Sorting Genetic Algorithm (NSGA-II) \cite{deb2002fast}, in which we define two competing objective functions: the likelihood of a solution (to be maximized), and the number of selected arcs (to be minimized). 
The aim of this work consists in investigating the behaviour of NSGA-II and in particular analyzing the quality of the solutions inferred by the method when solving the problem of learning the structure of a BN, \textcolor{black}{formulated as a multi-objective optimization problem}. 
To this regard, we thoroughly examined the results obtained with NSGA-II on a wide set of simulated data, by considering both the goodness of the inferred solutions in terms of the objective functions values achieved, and by comparing the retrieved structures with the ground truth, namely the models that were used to simulate the target data. 
We observed that NSGA-II can converge to solutions characterized by better likelihood and less arcs than classic Hill Climbing, although characterized in many cases by a lower similarity with the target network.

\section*{Materials and methods}
\label{sec:methods}

In this Section, we first report a formal definition of BNs and a description of the problem of learning their structure; then we describe the score-based and the population-based approaches, both in the case of single- and multi-objective optimization. 

\subsection*{Bayesian Networks and structure learning}
A Bayesian Network \cite{koller2009probabilistic} is a probabilistic graphical model describing a set of variables and their conditional dependencies by means of a directed acyclic graph (DAG). More precisely, let $G=(V,E)$ be a directed acyclic graph with nodes $V$ and arcs connecting the nodes $E$. 
The nodes in the DAG represent a set of random variables $\{v_1, \ldots, v_n\} \in V$. 
The structure of the DAG in turn induces a probability distribution over its nodes $\{v_1, \ldots, v_n\}$:
$$
\Probab{v_1, \ldots, v_n} = \prod_{i=1}^n \Pcond{v_i}{\pi_i}, \qquad
\Pcond{v_i}{\pi_i} = \bth_{v_i\mid \pi_i},
$$
where $\pi_i = \{ v_j \mid v_j \to v_i \in V\}$ are $v_i$'s parents in the DAG, and $\bth_{v_i\mid \pi_i}$ is a probability density function. 

Let now $O$ be a dataset of $m$ observations for the $n$ above mentioned random variables. Then, we can define the log-likelihood (\textcolor{black}{abbreviated as $LL$ in the rest of manuscript}) of the BN as: 
$$
LL(O|G) = \log\Pcond{O}{G,\bth} \, .
$$

However, such likelihood function is known to be monotonically increasing toward more complex solutions, that is, given an arbitrary network structure, adding arcs to it does not reduce its likelihood, hence leading to overfitting solutions \cite{koller2009probabilistic,caravagna2017learning}. 
To contain this problem when estimating the quality of a given network structure, the log likelihood of the network is usually coupled with a \textcolor{black}{regularizator} score $R(G)$ to penalize complex models over sparser ones: 
\begin{equation}
LL(O|G) = \log\Pcond{O}{G,\bth} - R(G)\, .
\label{eq:fitness}
\end{equation}
The \textcolor{black}{regularizator} term $R(G)$ in Eq. \ref{eq:fitness} is a penalty term for the number of parameters in the model and the size of the data. 
Specifically, two broadly used scores are the Akaike information criterion (AIC) \cite{akaike1974new} where $R(G) = |G|$ and the Bayesian information criterion (BIC) \cite{schwarz1978estimating} where $R(G) = \frac{|G|}{2}\log m$, where $m$ the number of samples. 

The problem of learning the structure of BNs can thus be formalized as an optimization problem where the goal is to maximize Eq. \ref{eq:fitness}, i.e., the likelihood of the data while minimizing its complexity with respect to the network's structure. 
This problem is known to be NP-complete \cite{chickering1996learning}, although greedy procedures such as hill climbing (HC) or taboo search on likelihood scores coupled with a \textcolor{black}{regularizator term have been shown to be surprisingly effective to solve it and have been widely adopted in real world applications \cite{koller2009probabilistic}}. 

The \textcolor{black}{regularizator} term is generally effective to obtain BNs characterized by high likelihood and a controlled number of arcs, but it may prevent meaningful solutions from being considered if the relative increment in the likelihood function when adding arcs is not high enough to counterbalance the penalization due to the \textcolor{black}{regularizator} term. 
In this paper we present an approach to explicitly account for model complexity when learning BNs by means of multi-objective optimization, where we simultaneously characterize high likelihood solutions (objective function $1$) and low model complexity (objective function $2$). 

\subsection*{Score-based approaches to learn the structure of BNs}

A classic approach to learn the structure of a BN formulates this task as an optimization problem consisting in the identification of a model that maximizes the likelihood of the data given the retrieved BN. 
However, the number of candidate parents per node (in-degree) greatly influences the efficacy in solving this task, in fact high in-degree implies a large search space and high computational time. 
Therefore, when the number of variables is big, the in-degree is generally constrained to be small in order to allow the optimization to be feasible. Conversely, when the number of variables is in the order of $10$ to $50$, one can allow for big  in-degree values. 
In this context, greedy search methods based on a likelihood score coupled with a \textcolor{black}{regularizator} term to account for model complexity, have been shown to be surprisingly effective in practice \cite{gamez2011learning,beretta2017quantitative,ramazzotti2017learning}. 

One of the simplest techniques that have been used to solve such optimization task is Hill Climbing (HC) \cite{koller2009probabilistic}. 
HC is based on the concept of neighborhood $N(i)$, defined for each valid solution $i$;  at each iteration, all solutions that are neighbor of the current one are evaluated, the next solution $j$ that leads to the best improvement is then selected among the solutions in $N(i)$. 
Formally, the neighborhood is a function $N:S\rightarrow 2^S$ that assigns to each solution in the search space $S$ a (non-empty) subset of $S$. 
In this context, $S$ comprises all the valid directed acyclic graphs and each solution $i$ is also a directed acyclic graph, which can be denoted as $G_i$. 

The steps characterizing the HC algorithm for a maximization problem of a given objective function (in our case the log-likelihood function $LL(O|G)$) are the following:
\begin{enumerate}
\item choose an initial solution $G_i$ in $S$;
\item find the best solution among those in $N(G_i)$;
\item if $LL(\textcolor{black}{O}|G_j) < LL(\textcolor{black}{O}|G_i)$, then stop; else set $G_i=G_j$ and return to Step 2.
\end{enumerate}

While being simple and intuitive, HC has been found to perform well in the context of struture learning for BNs \cite{gamez2011learning,beretta2017quantitative,ramazzotti2017learning}. For this reason, it represents a good baseline for comparing the performance of more advanced approaches.

\subsection*{Population-based optimization methods}

\paragraph*{Single-objective Optimization}

Genetic Algorithms (GAs) is a class of global search meta-heuristics inspired by the mechanisms of natural selection \cite{holland1975adaptation}. 
In GAs, a population ${\Pi}$ of candidate solutions (the individuals) iteratively evolves as a consequence of simulated selection, mutation and crossover mechanisms, converging to the global optimum with respect to a given fitness function $f$. 
 GAs were shown to be effective for BN learning, both in the case of available \cite{ramazzotti2016parallel} and not available \emph{a priori} knowledge about \textcolor{black}{a partially ordered set involving the nodes of the network, i.e., for certain pairs nodes, one of them precedes the other in the ordering leading to a directed acyclic graph}\cite{larranaga1996structure}. 
In the context of BNs inference, the population ${\Pi}$ is usually composed of $Q$ randomly created binary strings, \textcolor{black}{i.e., ordered sequences of binary digits that represent linearized versions of the adjacency} matrices of candidate BNs with $K$ nodes.
The individuals in ${\Pi}$ undergo an iterative process whereby three genetic operators, i.e., selection, crossover and mutation, are applied in sequence to simulate the evolution process, which results in a new population of possibly improved solutions. 
\textcolor{black}{We hereby briefly summarize the three genetic operators:
\begin{itemize}
    \item during the selection process, $Q$ the individuals from ${\Pi}$ are chosen with replacement, with a probability that is proportional to the fitness value \cite{back1994selective}; 
    \item the selected individuals then undergo the crossover operator, with a user-defined probability $p_{\chi}$. During this process, the structure of the individuals is recombined so that promising parents can possibly yield improved offspring;
    \item finally, the mutation operator is used to introduce new genetic materials in the population allowing for a further exploration of the search space. The mutation operator traditionally flips an arbitrary bit of the individual, with a probability $p_{\mu}$.
\end{itemize}
}

It is worth noting that, in the case of ordered nodes, both  crossover and mutation are \emph{closed} operators, because the resulting offspring always encode valid DAGs. 
To the aim of ensuring a consistent population of individuals throughout the generations, in the case of unordered nodes, the two operators are followed by a correction procedure, in which the candidate BN is analyzed to identify the  presence of invalid cycles. 
Alternative representations of BNs \cite{de2017novel}, tailored for evolutionary methods and designed to prevent the creation of cycles in the graph as a result of the genetic operators, have been proposed in the literature but, for the sake of simplicity, they will not be considered in this paper. 
For further information about the correction phase exploited in this work, we refer the interested reader to \cite{ramazzotti2016parallel}. 

\paragraph{Multi-objective Optimization}
GAs aim at the identification of the global optimum of a given fitness function. 
For this reason, they are not suitable for the simultaneous optimization of multiple (possibly opposing) criteria $f_1, f_2, \dots, f_\Omega$, where $\Omega \geq 2$. 
This might be the case of BNs inference, where the goal is to maximize the likelihood ($f_1$) while keeping the connectivity of the network as reduced as possible ($f_2$). 
Although regularizators such as AIC or BIC provide powerful heuristics to reduce the BN inference to a single-objective problem, in this paper we investigate the effectiveness of a radically different approach based on evolutionary algorithms able to support multi-objective optimization (MOO) \cite{miettinen2012nonlinear}. 
This class of algorithms aims at the identification of the Pareto front of non-dominated solutions, that is, the set of optimal solutions that cannot be further improved without affecting one of the fitness values. 
Formally, MOO algorithms are based on the notion of domination: an individual $y_1 \in {\Pi}$  \emph{dominates}\footnote{In this case, we assume that all objectives must be maximized. This formalization of domination can be straightforwardly modified to consider the case of the  minimization of (a subset of) the objectives.} another individual $y_2 \in {\Pi}$ if: 

\begin{itemize}
\item $f_i(y_1) \geq f_i(y_2)$ for all $i=1, \dots, \Omega$;
\item $f_j(y_1) > f_j(y_2)$ for at least one index $j$ in $1, \dots, \Omega$.
\end{itemize}

Given an arbitrary population, a ranking of non-dominated solutions can be created using the following procedure: 

\begin{itemize}
\item a variable $s \in \mathbb{N}$ is initialized to $1$; 
\item all individuals belonging to the non-dominated front are identified and copied from ${\Pi}$ to a set ${D}_s$; 
\item ${\Pi}$ is replaced by ${\Pi} \setminus {D}_s$;
\item $s$ is incremented by 1;
\item the process iterates until ${\Pi}=\varnothing$. 
\end{itemize}

The ranking of dominated/dominating  solutions is essential for the functioning of  MOO algorithms. 

In the context of MOO, the most widespread methodology is the Non-dominated Sorting Genetic Algorithm (NSGA-II) \cite{deb2002fast}.
NGSA-II is an elitist method, notably characterized by a computational complexity of $\mathcal{O}(\Omega Q^2)$, where $\Omega$ is the number of objectives  and $Q$ is the population size. 

NSGA-II's functioning can be sketched as follows. 
The algorithm starts by randomly generating a population ${\Pi}$ of \textcolor{black}{$Q$} individuals and a new offspring population ${\Pi}'$. 
Then, since  NGSA-II is an elitist method, it  creates a temporary population ${U} = {\Pi} \cup {\Pi}'$. 
The individuals belonging to the best non-dominated sets in ${U}$ are progressively inserted into a novel population ${\Pi}''$, until exactly \textcolor{black}{$Q$} individuals are selected. 
During this process, however, the $j$-th non-dominated sets ${D}_j$ might have more individuals than necessary (i.e., $|{\Pi}''| + |{D}_j| > $\textcolor{black}{$Q$}). 
In such a case, NSGA-II  creates a further ranking of the individuals in ${D}_j$ using a crowded-comparison operator $\prec_n$, which calculates the crowding distance of each putative solution \cite{deb2002fast}.
By using the crowding ranking, NGSA-II deterministically selects the most ``diverse'' solutions, trying to maintain a high level of diversity in the population, and completes the new population. 
At the end of the selection process, NSGA-II executes crossover and mutation on the individuals of ${\Pi}''$; in particular, here we exploit the genetic operators of GAs described in the previous section.
Similarly to other evolutionary methods, NSGA-II iterates until a halting  criterion is verified, e.g., after a fixed number of generations $IT_{\texttt{max}}$. 

To solve the problem of learning the structure of a BN, we exploit NSGA-II to optimize $\Omega = 2$ objectives:
\begin{itemize}
	\item the first objective is to maximize the likelihood function defined in Eq. \ref{eq:fitness};
	\item the second (conflicting) objective is to minimize the number of arcs in the Bayesian Network.
\end{itemize}

There exist alternative methods for MOO in the literature, 
notably the Strength Pareto Evolutionary Algorithm (SPEA) \cite{zitzler1999multiobjective}, which performs MOO by assigning a fitness value to individuals that is proportional to the number solutions they dominate; moreover, SPEA leverages a clustering procedure to contain the size of the nondominated archive. 
In the SPEA2 variant \cite{zitzler2001spea2}, an alternative fitness evaluation scheme is exploited,  a density estimation  method is used to improve the search in the case of MOO with more than $2$ objectives,  and an improved truncation scheme is adopted to preserve the solutions on the extreme sides of the Pareto front approximation.
Further examples of MOO algorithms are multi-objective Particle Swarm Optimization (MOPSO) \cite{coello2002mopso} and its variants \cite{nebro2009smpso,britto2012mopso} are MOO methods specifically defined for problems whose candidate solutions can be encoded as fixed-length vectors of real values.
Although these algorithms could, in principle, be adapted to deal with discrete candidate solutions (as in the case of BN inference), we considered NSGA-II as a more natural and straightforward choice for the data structures we need to evolve.
For the same reason, we did not consider the multi-objective formulations of CMA-ES \cite{igel2007covariance}.
Finally, an extended version of NSGA, called NSGA-III, was also published \cite{deb2013evolutionary}.
NSGA-III is a novel reference-point-based evolutionary algorithm for many-objective optimization (i.e., characterized by 4 or more objectives); we did not consider this method in this work on account of its additional complexity and the fact that we are optimizing only 2 objectives.

\section*{Results}
\label{sec:results}
In this work, we represent the candidate BNs as adjacency matrices encoded by means of binary strings with fixed length, as described in \cite{ramazzotti2016parallel}. 
Thus, the search space explored by the evolutionary algorithms consists in all possible configurations of bits representing valid DAGs.
The NSGA-II settings used for the tests reported in this section are listed in Table \ref{tab:nsga}.
\textcolor{black}{
In all tests that follow we used the following python libraries: DEAP \cite{DEAP_JMLR2012}, NumPy \cite{harris2020array}, NetworkX \cite{SciPyProceedings_11}. 
The likelihood calculations and the HC were performed using the bnlearn R library \cite{bnlearn}.
The source code of the methods employed in this work is available for download at the following address: \url{https://github.com/aresio/moo-bn}.}

\begin{table}[tb]
	\centering
	\caption{NSGA-II settings used in this work.}
	\label{tab:nsga}
	\begin{tabular}{l|cc}
	\hline
	\textbf{Setting} & \textbf{Value}  \\
	\hline
	$p_{\chi}$ & $0.90$ \\
	$p_{\mu}$ & $0.05$ \\
	$Q$ & $128$ \\
	$IT_{\texttt{max}}$ & $100$ \\
	
	\hline
	\end{tabular}
\end{table}

In order to compare the performance of BN learning using NSGA-II with the classic HC combined with BIC or AIC \textcolor{black}{regularizator}---which is typically exploited to learn BN structures \cite{koller2009probabilistic}---we generated a set of networks with $n=10$ variables and random (different) structural characteristics.
The probability distributions induced by such networks were exploited to generate multiple datasets characterized by an increasing number of samples ($m=50, 100, 500$) for the observed variables, which were later used to assess the log-likelihood during the inference processes (i.e., fitness evaluations). 
For each dataset, we performed multiple tests with an increasing level of noise (i.e., random flipping in the observation matrices) to model any potential source of errors that can naturally occur during an experimental data collection; specifically, we considered a noise level equal to $0\%$, $10\%$, $20\%$.
Finally, we employed $3$ different levels of density ($0.2$, $0.5$, $0.8$), i.e., number of edges with respect to number of variables.

We obtained a total of $27$ different scenarios by combining the sample size, density and noise levels; for each scenario, we performed $50$ independent repetitions of the BN inference with HC and NSGA-II, in order to collect statistically significant results.

For the sake of clarity, we report here only a subset of the results but we describe the emerging trends observed by considering all the obtained results.
As a matter of fact, the results concerning the density level equal to $0.5$ are not shown as they are similar to those achieved in the case of density level equal to $0.2$.
\textcolor{black}{The complete set of results obtained from the tests performed in this work is provided in the Supplementary File.}
In each figure we report a scatterplot (left panel) in which we show all Pareto fronts obtained at the end of each NSGA-II run (orange circles), and we highlight the optimal Pareto front (orange solid circles); the solutions identified by BIC are denoted by a blue square; the solutions identified by AIC are denoted by a purple diamond; the ground truth (the model used to generate the target data) is shown as a black star.
In the right \textcolor{black}{panel} of each figure we show the violin plots describing the distributions of precision, recall and specificity values obtained by HC with AIC and BIC \textcolor{black}{regularizator}, and NSGA-II considering the bests Pareto front according to log-likelihood (top LL) and to the \textcolor{black}{regularizator} score (top reg), and the median Pareto front obtained by sorting all solutions with respect to log-likelihood values (median PF).
Precision, recall and specificity are calculated as described in \cite{powers2011evaluation,beretta2017quantitative}.

\begin{figure*}[!ht]
  \centering
 \begin{minipage}{.49\linewidth}
  \includegraphics[width=\textwidth,trim=0 0 0 25, clip]{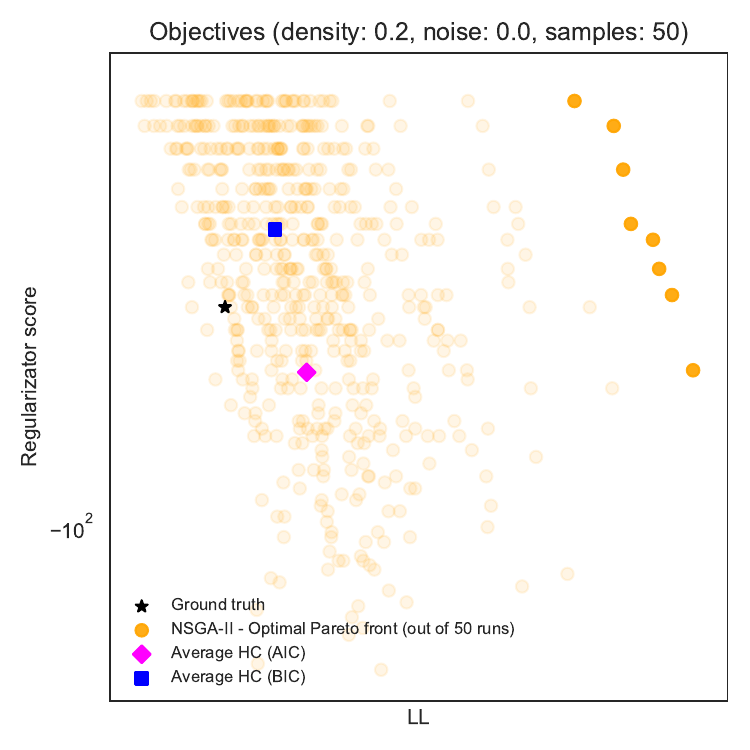}
  \end{minipage}
  \begin{minipage}{.5\linewidth}
  \includegraphics[width=\textwidth]{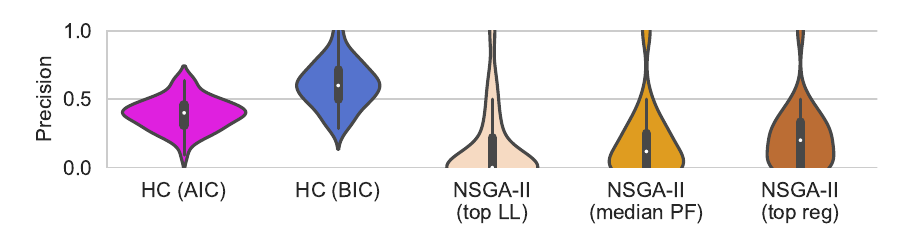}
   \includegraphics[width=\textwidth]{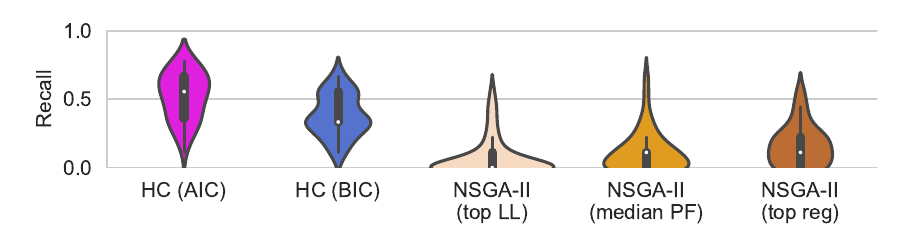}
   \includegraphics[width=\textwidth]{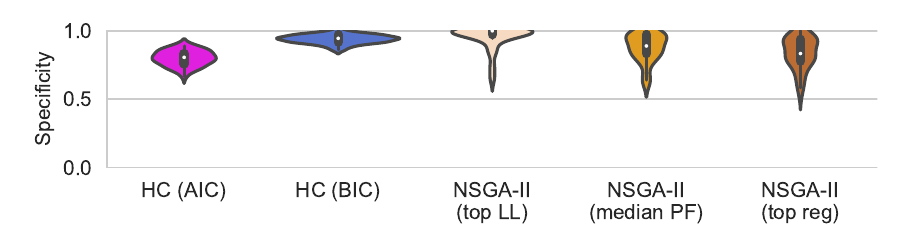}
  \end{minipage}
  \caption{Results of the BN learning in the case of networks with density equal to $0.2$, $50$ samples and $0\%$ noise.
  Comparison of the objective functions (left), and distributions of the values of precision, recall and specificity (right) of the solutions. The solutions found by NSGA-II dominate those found by HC. However, the solutions found by NSGA-II are characterized by very low precision and recall, although they have very high specificity.  }
  \label{fig:figure1}
\end{figure*}

\begin{figure*}[!ht]
\centering
   \begin{minipage}{.49\linewidth}
  \includegraphics[width=\textwidth,trim=0 0 0 25, clip]{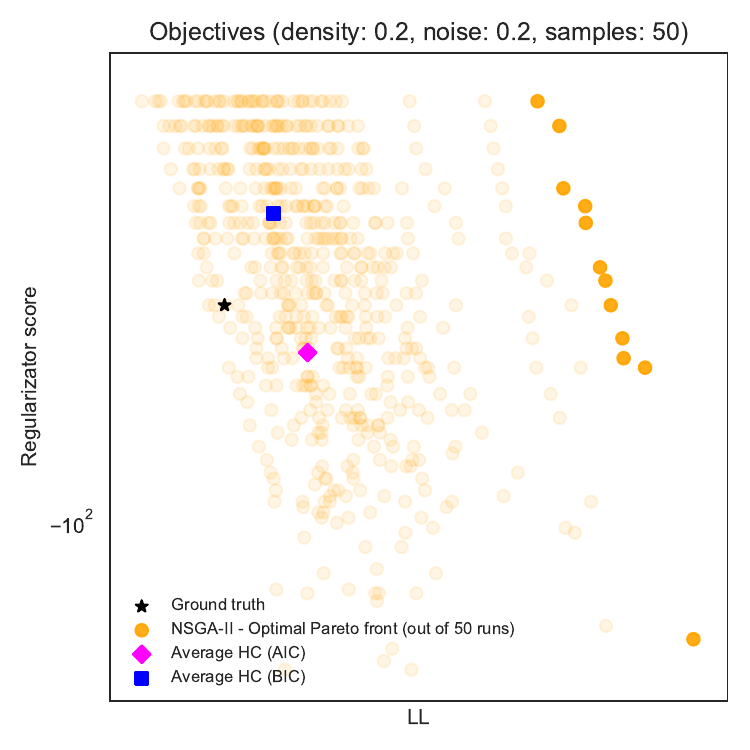}
  \end{minipage}
  \begin{minipage}{.5\linewidth}
    \includegraphics[width=\textwidth]{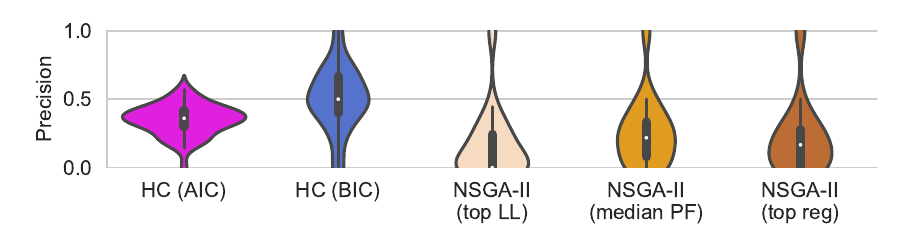}
   \includegraphics[width=\textwidth]{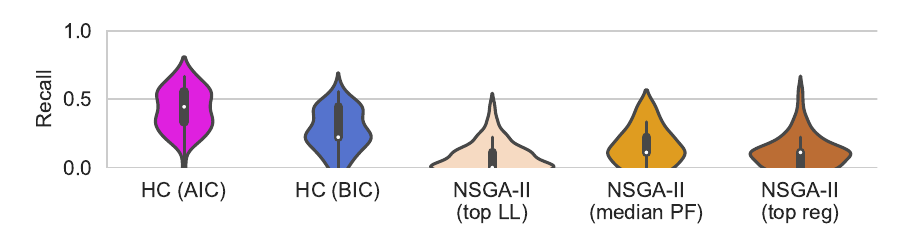}
   \includegraphics[width=\textwidth]{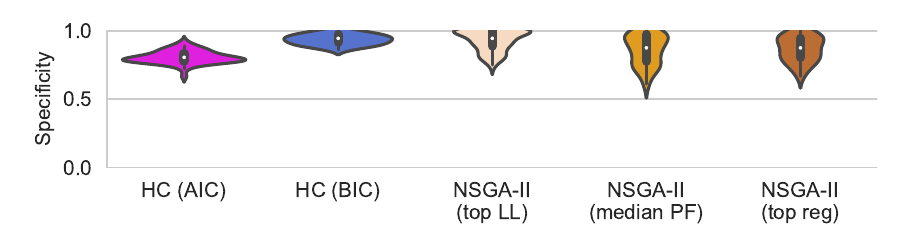}
  \end{minipage}
    \caption{Results of the BN learning in the case of networks with density equal to $0.2$, $50$ samples and $20\%$ noise.
    Comparison of the objective functions (left), and distributions of the values of precision, recall and specificity (right) of the solutions.  The solutions found by NSGA-II dominate those found by HC. Moreover, with such high density value, the solutions found by NSGA-II are characterized by high precision and specificity, with performances similar to HC.
    }
  \label{fig:figure2}
\end{figure*}

\begin{figure*}[!ht]
\centering
  \begin{minipage}{.49\linewidth}
  \includegraphics[width=\textwidth,trim=0 0 0 25, clip]{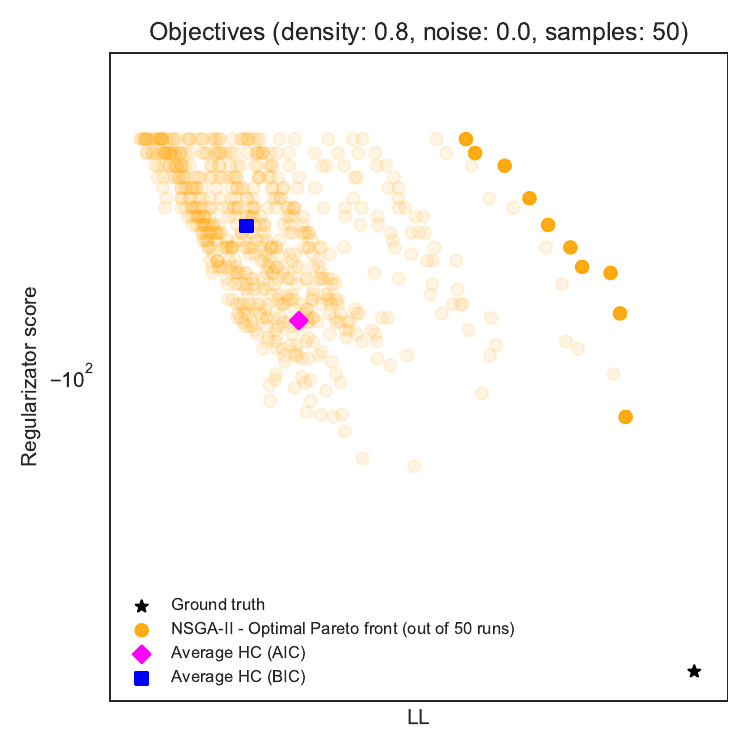}
  \end{minipage}
  \begin{minipage}{.5\linewidth}
  \includegraphics[width=\textwidth]{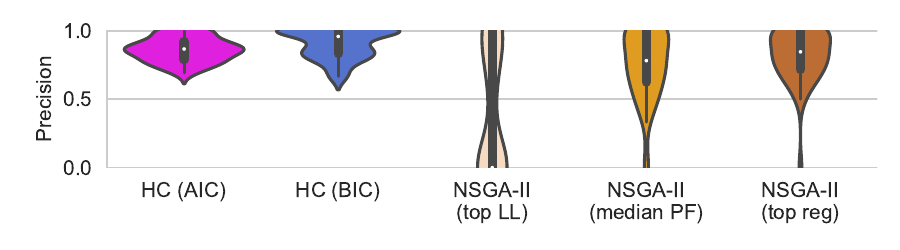}
  \includegraphics[width=\textwidth]{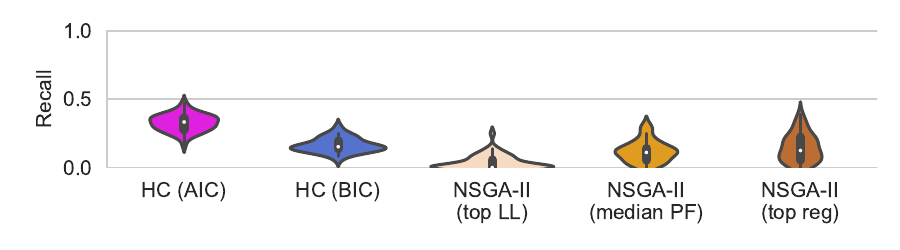}
    \includegraphics[width=\textwidth]{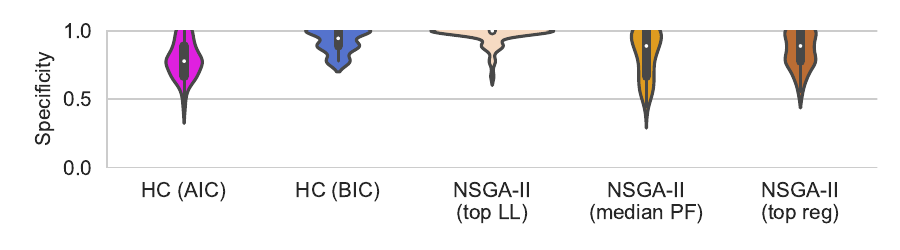}
  \end{minipage}
   
  \caption{Results of the BN learning in the case of networks with density equal to $0.8$, $50$ samples and $0\%$ noise.
  Comparison of the objective functions (left), and distributions of the values of precision, recall and specificity (right) of the solutions.  The solutions found by NSGA-II dominate those found by HC. Moreover, with such high density value, the solutions found by NSGA-II are characterized by high precision and specificity, with performances similar to HC.}
  \label{fig:figure3}
\end{figure*}

\begin{figure*}[!ht]
\centering
  \begin{minipage}{.49\linewidth}
  \includegraphics[width=\textwidth,trim=0 0 0 25, clip]{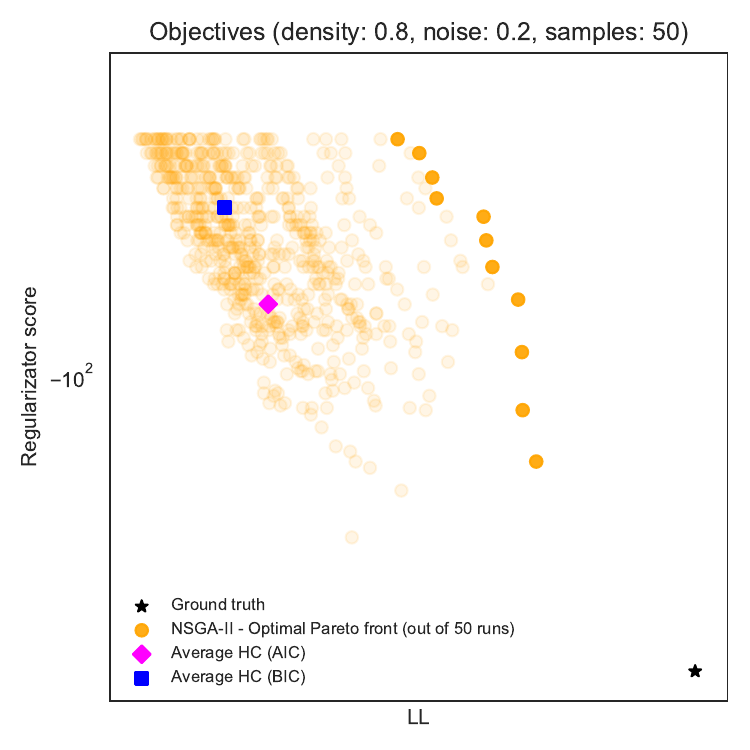}
  \end{minipage}
  \begin{minipage}{.5\linewidth}
   \includegraphics[width=\textwidth]{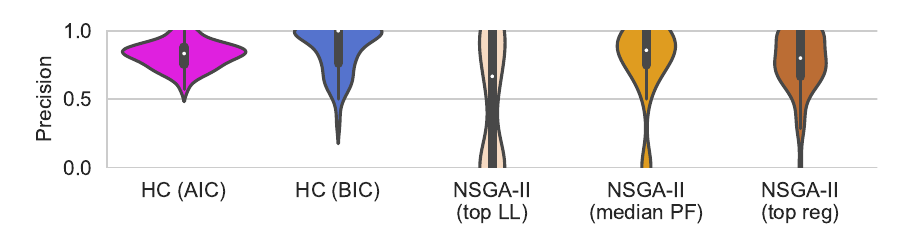}
   \includegraphics[width=\textwidth]{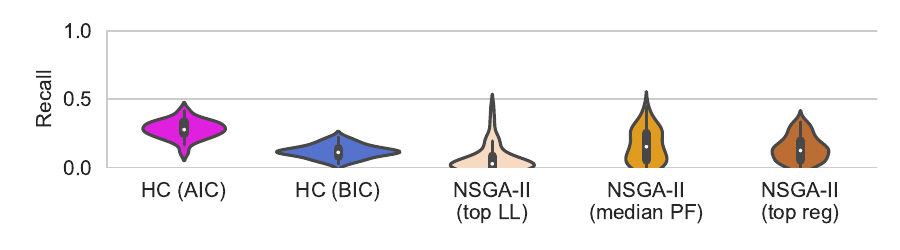}
   \includegraphics[width=\textwidth]{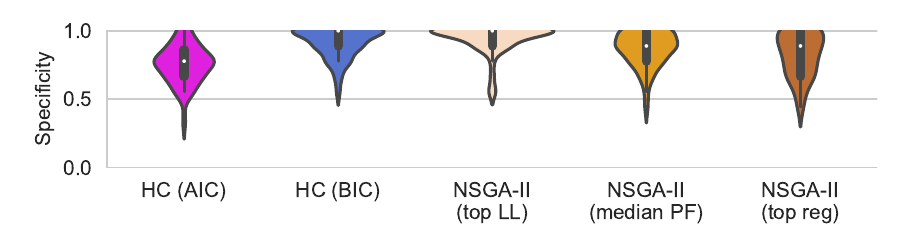}
  \end{minipage}
  \caption{Results of the BN learning in the case of networks with density equal to $0.8$, $50$ samples and $20\%$ noise.
  Comparison of the objective functions (left), and distributions of the values of precision, recall and specificity (right) of the solutions. The solutions found by NSGA-II dominate those found by HC. Moreover, with such high density value, the solutions found by NSGA-II are characterized by high precision and specificity, with performances similar to HC.}\label{fig:figure4}
\end{figure*}

\begin{figure*}[!ht]
\centering
  \begin{minipage}{.49\linewidth}
  \includegraphics[width=\textwidth,trim=0 0 0 25, clip]{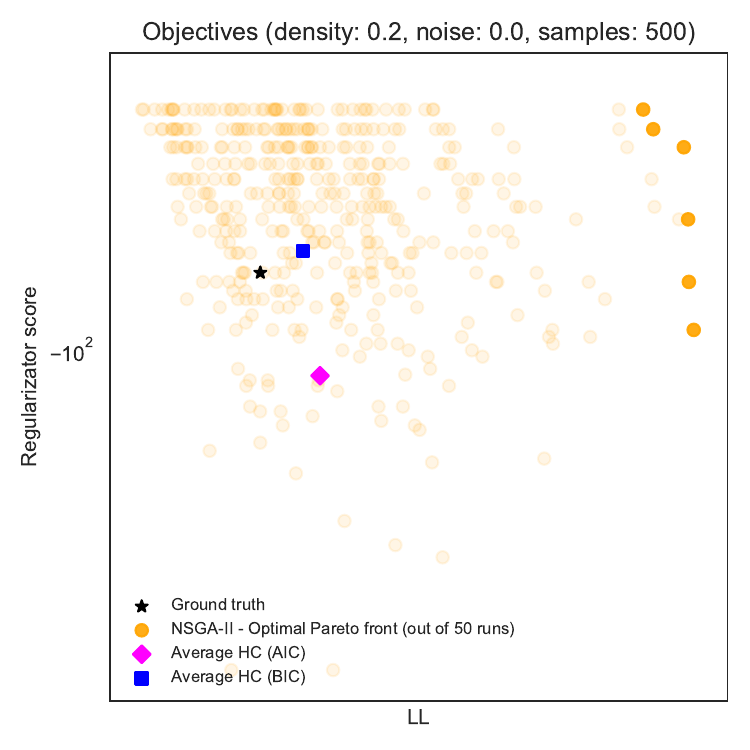}
  \end{minipage}
  \begin{minipage}{.5\linewidth}
  \includegraphics[width=\textwidth]{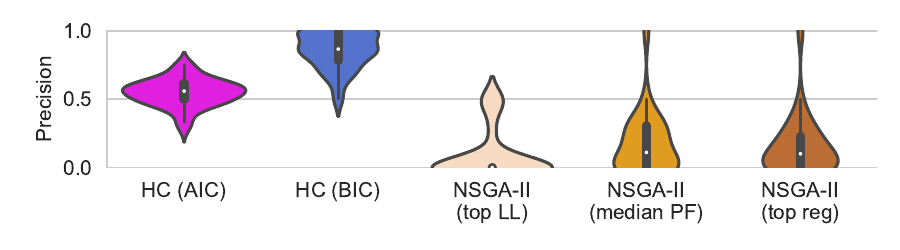}
   \includegraphics[width=\textwidth]{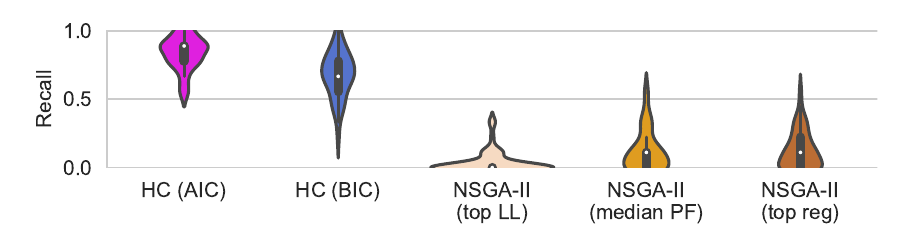}
   \includegraphics[width=\textwidth]{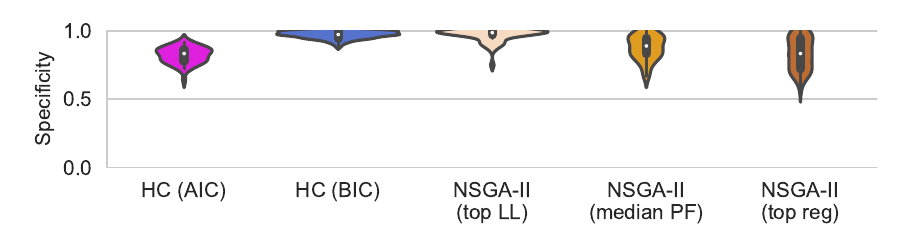}
  \end{minipage}
  \caption{Results of the BN learning in the case of networks with density equal to $0.2$, $500$ samples and $0\%$ noise.
  Comparison of the objective functions (left), and distributions of the values of precision, recall and specificity (right) of the solutions. It is worth noting that both approaches (NSGA-II and HC) lead to solutions characterized by better likelihood, and less arcs, than the ground truth.}\label{fig:figure5}
\end{figure*}

\begin{figure*}[!ht]
  \centering
  \begin{minipage}{.49\linewidth}
  \includegraphics[width=\textwidth,trim=0 0 0 25, clip]{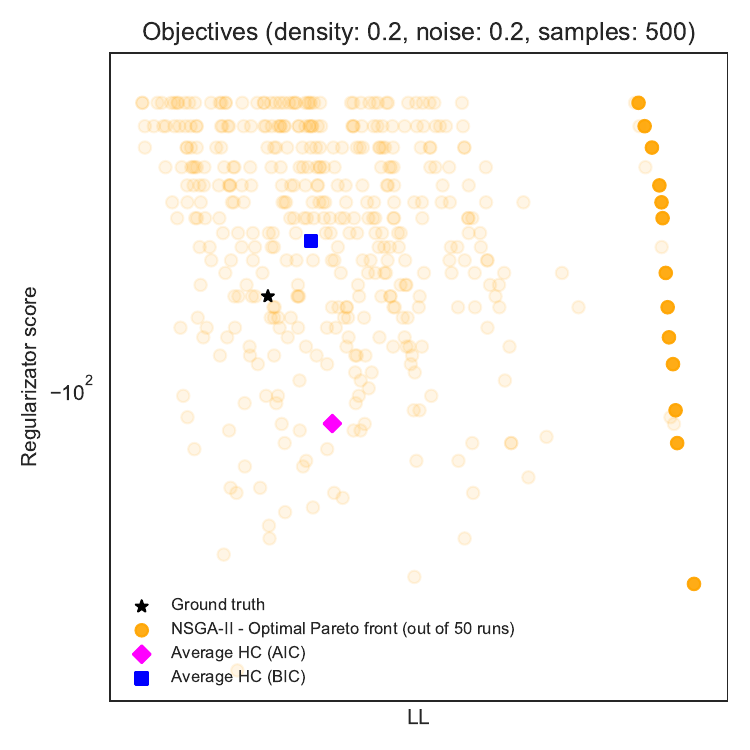}
  \end{minipage}
  \begin{minipage}{.5\linewidth}
  \includegraphics[width=\textwidth]{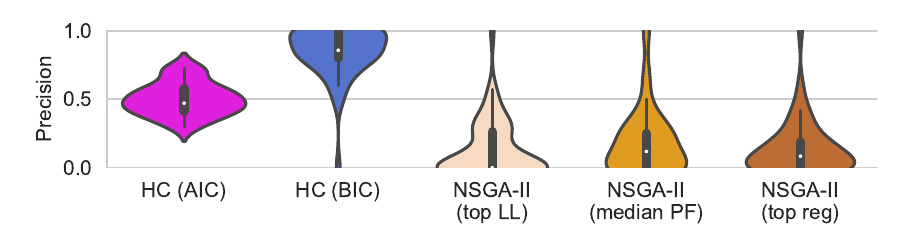}
   \includegraphics[width=\textwidth]{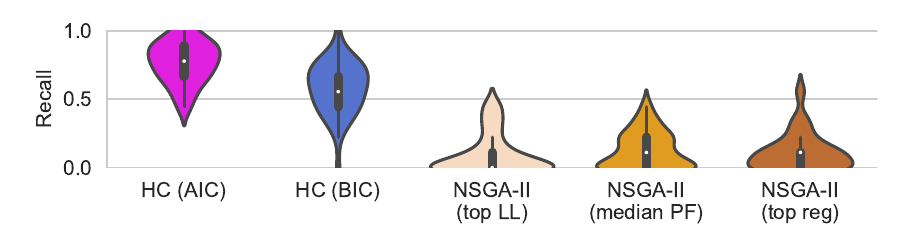}
   \includegraphics[width=\textwidth]{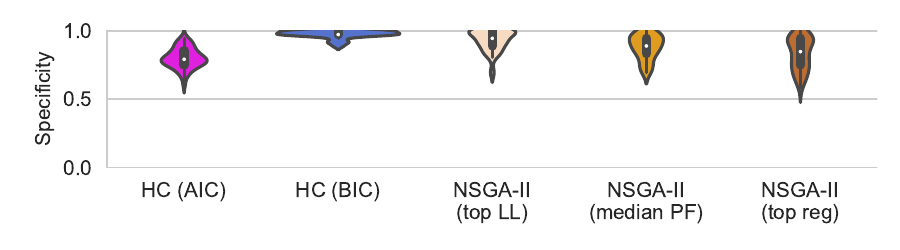}
  \end{minipage}
  \caption{Results of the BN learning in the case of networks with density equal to $0.2$, $500$ samples and $20\%$ noise.
  Comparison of the objective functions (left), and distributions of the values of precision, recall and specificity (right) of the solutions. It is worth noting that both approaches (NSGA-II and HC) lead to solutions characterized by better likelihood, and less arcs, than the ground truth. Compared to Figure \ref{fig:figure5}, noise does not seem to affect NSGA-II's performances.}\label{fig:figure6}
\end{figure*}

\begin{figure*}[!ht]
  \centering
   \begin{minipage}{.49\linewidth}
  \includegraphics[width=\textwidth,trim=0 0 0 25, clip]{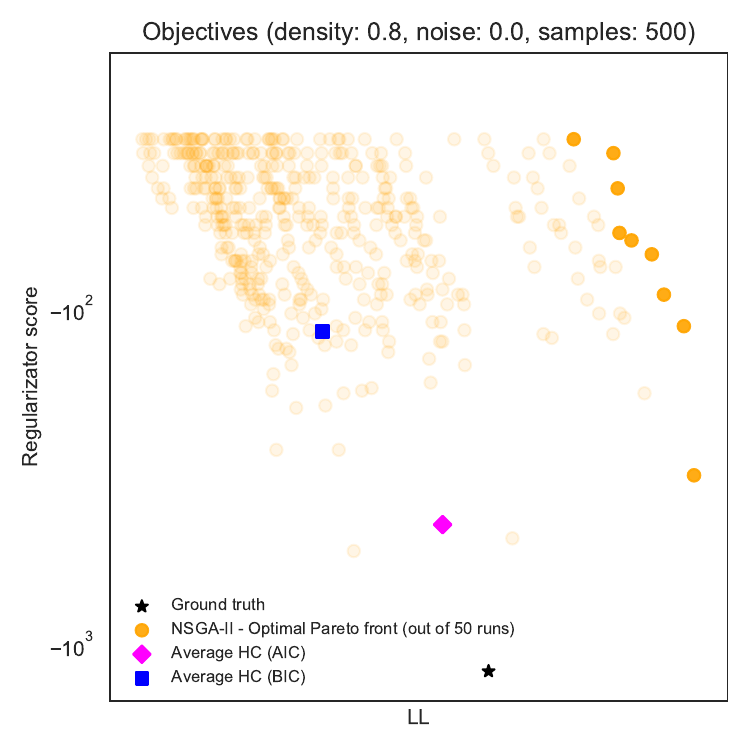}
  \end{minipage}
  \begin{minipage}{.5\linewidth}
    \includegraphics[width=\textwidth]{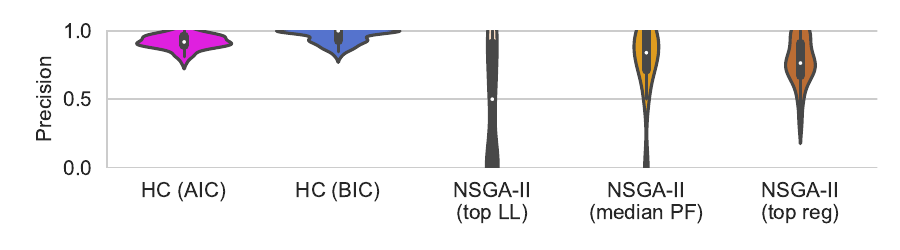}
   \includegraphics[width=\textwidth]{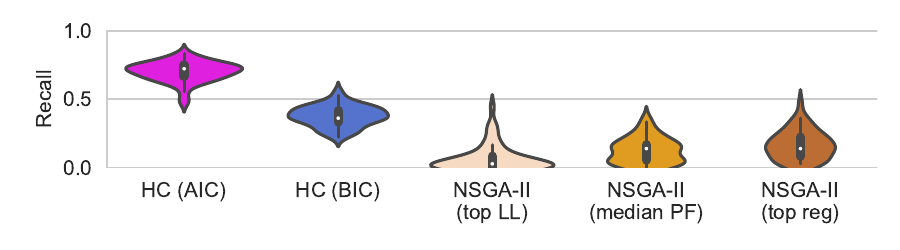}
   \includegraphics[width=\textwidth]{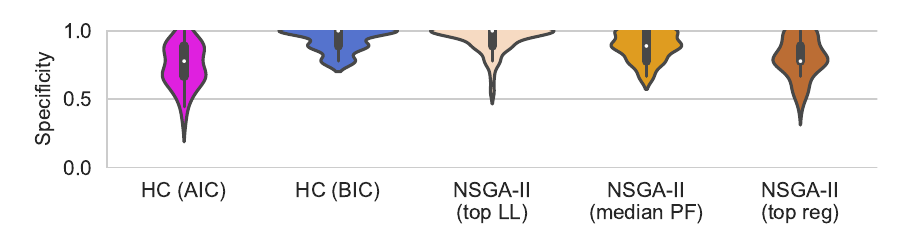}
  \end{minipage}
  \caption{Results of the BN learning in the case of networks with density equal to $0.8$, $500$ samples and $0\%$ noise.
  Comparison of the objective functions (left), and distributions of the values of precision, recall and specificity (right) of the solutions.
  }\label{fig:figure7}
\end{figure*}

\begin{figure*}[!ht]
 \centering
   \begin{minipage}{.49\linewidth}
  \includegraphics[width=\textwidth,trim=0 0 0 25, clip]{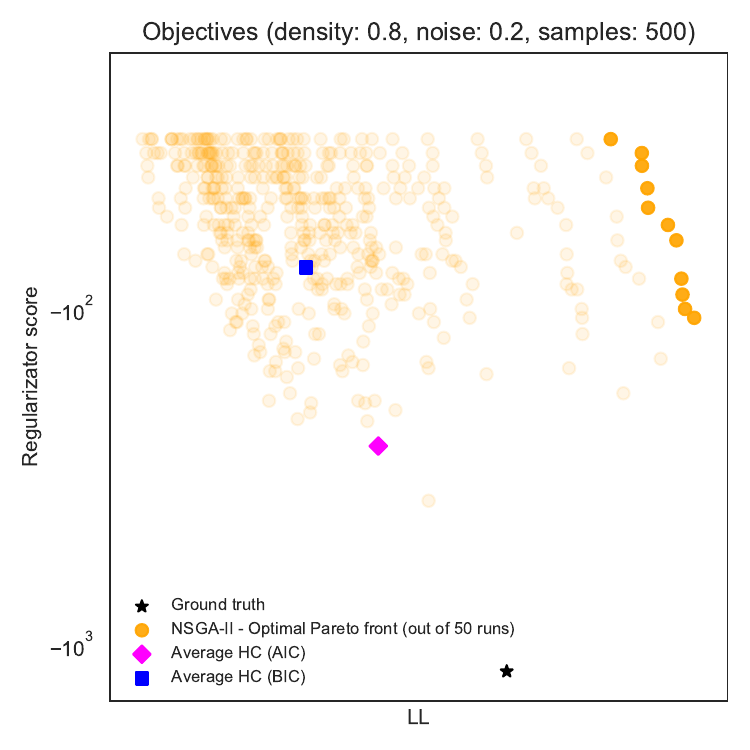}
  \end{minipage}
  \begin{minipage}{.5\linewidth}
    \includegraphics[width=\textwidth]{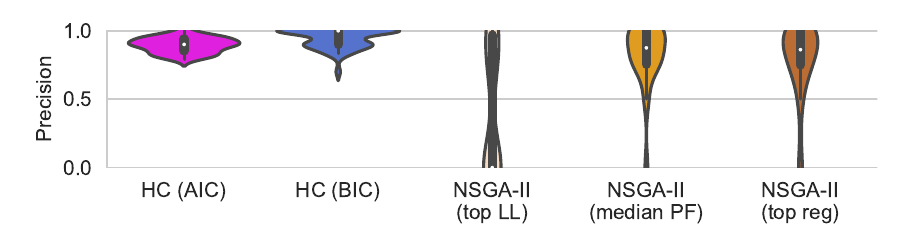}
   \includegraphics[width=\textwidth]{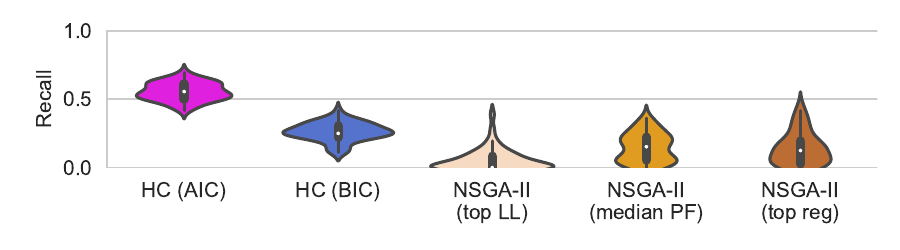}
   \includegraphics[width=\textwidth]{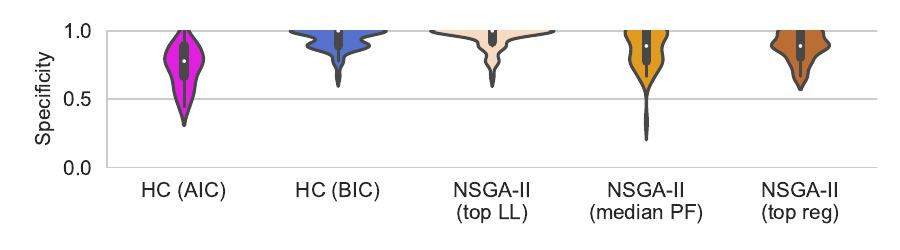}
  \end{minipage}
  \caption{Results of the BN learning in the case of networks with density equal to $0.8$, $500$ samples and $20\%$ noise.
  Comparison of the objective functions (left), and distributions of the values of precision, recall and specificity (right) of the solutions. Compared to Figure \ref{fig:figure7}, noise does not seem to affect NSGA-II's performances, while affecting HC.}\label{fig:figure8}
\end{figure*}

In Figure \ref{fig:figure1} (left panel) we show the comparison of the Pareto fronts produced by NSGA-II against BIC and AIC, in the case of networks characterized by a low number of edges (density equal to $0.2$), by using a dataset with $50$ samples and no noise.
This result shows that the optimal Pareto front largely dominate the solutions found by both BIC and AIC. 
It is worth noting that all algorithms converged to solutions having a log-likelihood higher than the ground truth.

Even though NSGA-II outperforms the other methods---from the point of view of objective functions optimization---the statistical analysis of the structural features obtained by considering precision, recall and specificity (Figure \ref{fig:figure1}, right), shows that AIC and BIC lead to better fitting solutions.
In particular, the solutions of the Pareto front are characterized by lower precision and recall values with respect to those identified by BIC and AIC, while the specificity of all solutions is comparable.
From this perspective, BIC leads to higher precision and specificity.
Due to the pressure introduced by the BIC regularizator, HC iteratively  determines the edges leading to the highest increment in the likelihood score, and adds those edges to the BN.
Thus, this heuristic seems to be more effective in identifying correct edges, even though it can affect the overall recall.
On the contrary, NSGA-II proceeds by random mutations, so that groups of edges---which would be discarded by HC as they slightly affect the overall likelihood score if added one by one to a solution---can be nevertheless added to the candidate solution. 
These erroneous edges (with respect to the unknown underlying BN), can increase the likelihood ($f_1$) and sometimes also lead to lower values of $f_2$ even with respect to the ground truth.
Similar results have been achieved in the case of density equal to $0.2$, $50$ samples and $10\%$ noise \textcolor{black}{(see Supplementary File)}.

Figure \ref{fig:figure2} shows the results of the tests executed in the case of a noise level equal to $20\%$ in the dataset.
Interestingly, all algorithms performed similarly to the previous case, showing their robustness to noisy datasets, in this scenario.

Figure \ref{fig:figure3} shows the results concerning highly connected BNs (density level $0.8$), in the case of datasets with $50$ samples and no noise.
In this condition, the performance of the three algorithms are similar, even though the optimal Pareto front identified by NSGA-II dominates the solutions found by AIC and BIC (left). 
It is worth noting that all algorithms identified solutions with fewer edges than the ground truth, while only the solutions of the optimal Pareto front of NSGA-II have \textcolor{black}{likelihood values similar to the ground truth}.
Moreover, NSGA-II obtained good results in terms of precision and specificity, reaching the levels of AIC and BIC (Figure \ref{fig:figure3}, right).

The same observations hold when $20\%$ noise is added to the dataset, where the only difference with the previous results regards the solution identified by AIC, whose precision and specificity are slightly lower (see Figure \ref{fig:figure4}).

The following set of tests were performed by considering $100$ (\textcolor{black}{see Supplementary File}) and $500$ samples, with the different values of density and noise in the datasets listed above.
Figures \ref{fig:figure5} and \ref{fig:figure6} show the results obtained in the case of $500$ samples, density $0.2$ and different values of noise in the datasets ($0\%$ and $20\%$, respectively).
As in the case of $50$ samples, the Pareto front largely dominates the solutions found by both BIC and AIC (left panels); moreover, we still observe that the statistical analysis of the structural features obtained by considering precision, recall and specificity (right panels of the figures),  shows that AIC and BIC lead to better fitting solutions than NSGA-II.

Figures \ref{fig:figure7} and \ref{fig:figure8} show the results obtained when considering a network density equal to $0.8$.
In general, the results achieved in \textcolor{black}{this batch of} tests are similar to those obtained in the case of datasets with $50$ samples, with a slight improvement of the performance of NSGA-II.

\begin{figure*}[!ht]
  \centering
  \includegraphics[width=.6\textwidth,trim=0 0 0 25, clip]{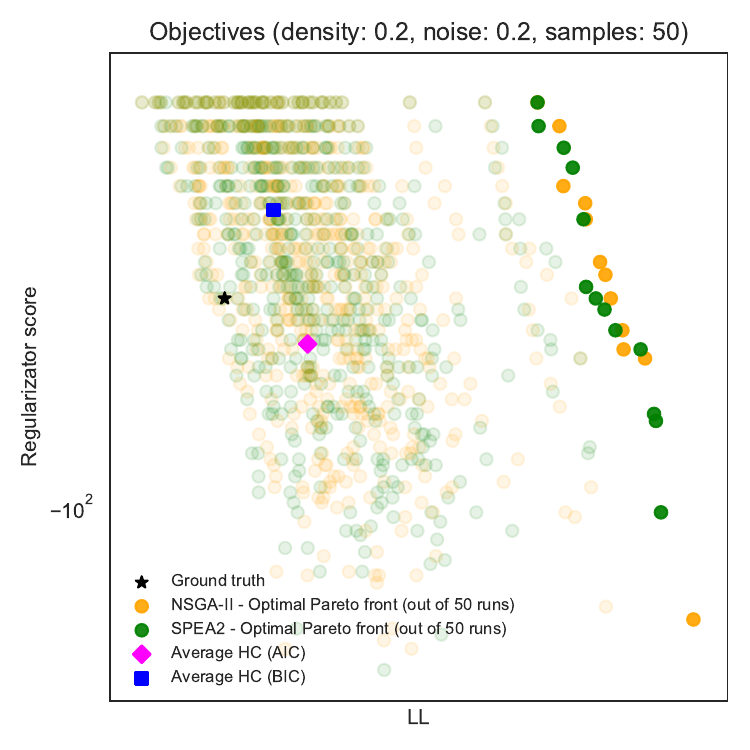}
  \caption{Results of the BN learning in the case of networks with density equal to $0.2$, $50$ samples and $20\%$ noise.
  Pareto fronts obtained at the end of each run are of NSGA-II (SPEA2) are denoted by orange (green) circles. The solution identified by HC with AIC (BIC) is denoted by a purple diamond (blue square), and the ground truth is indicated as a black star.}\label{fig:spea2}
\end{figure*}

As an additional test, we analyzed \textcolor{black}{and compared} the performance of NSGA-II and SPEA2, to verify that all results and observations concerning the comparison between HC and MOO were independent from the specific algorithm used.
As can be noted from Figure \ref{fig:spea2}, NSGA-II and SPEA2 solutions are characterized by the same values of both objective functions, and the optimal Pareto fronts dominate both the HC solutions and the ground truth\textcolor{black}{, corroborating the observations reported above}.

\begin{figure*}[!ht]
  \centering
  \includegraphics[width=.90\textwidth]{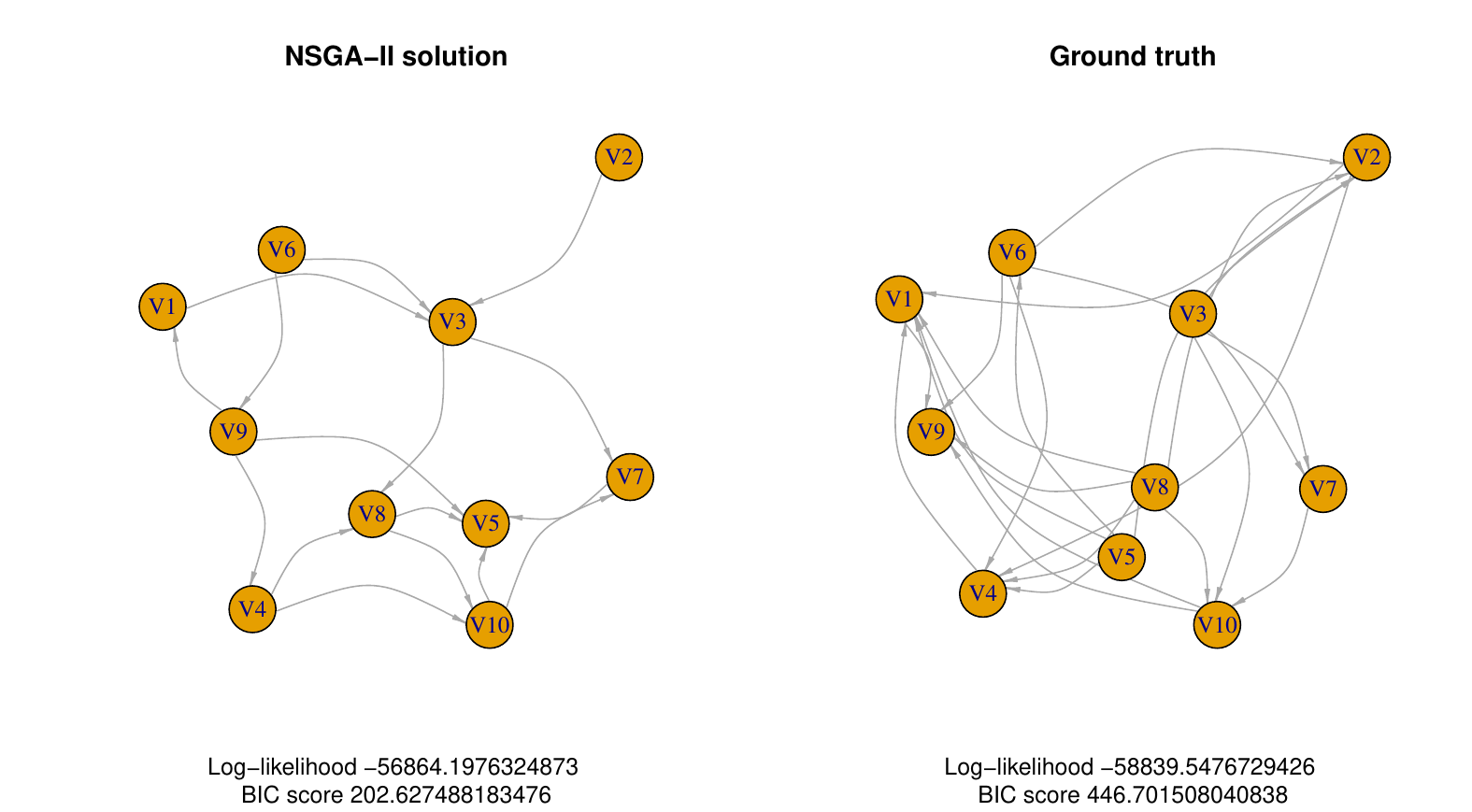}
  \caption{Comparison of a  BN identified by NSGA-II (on the left) against the ground truth (on the right). The network optimized by the evolutionary algorithm is characterized by less edges (16 instead of 23) and better likelihood. Hence, it represents a dominating solution even with respect to the ground truth.}
  \label{fig:comparison}
\end{figure*}

\begin{figure*}[!ht]
    \centering
    \includegraphics[width=.60\textwidth]{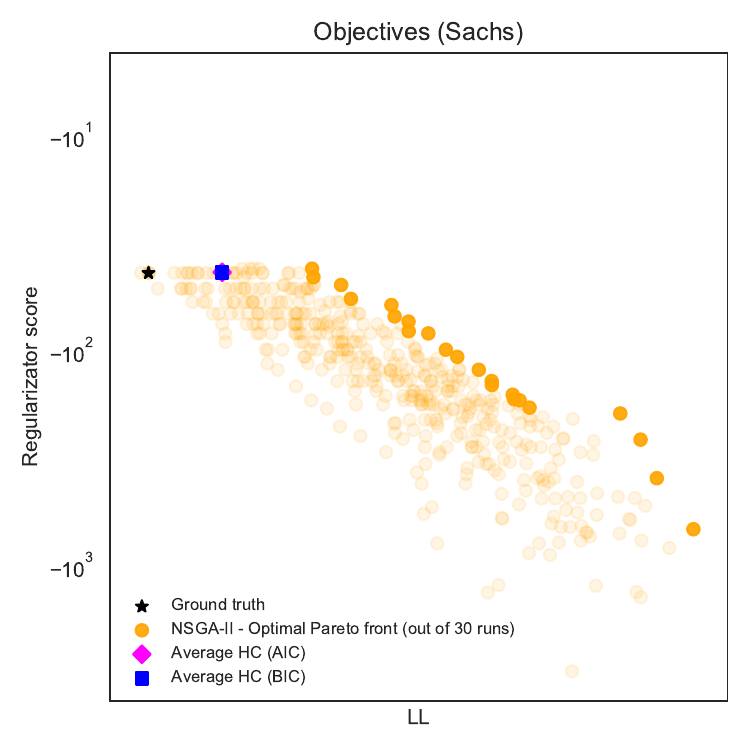}
    \caption{Results of the BN learning in the case of the signaling model presented by Sachs \emph{et al}. \cite{sachs2005causal}. Pareto fronts obtained at the end of each run are of NSGA-II are denoted by orange circles. The solution identified by HC with AIC (BIC) is denoted by a purple diamond (blue square), and the ground truth is indicated as a black star.}
    \label{fig:sachs}
\end{figure*}

In order to provide an example of the peculiarities of the  dominant solutions found by \textcolor{black}{the MOO algorithm}, we show in Figure \ref{fig:comparison} a comparison of the structures of an optimal BN produced by NSGA-II (left) that completely dominates the ground truth (right).
The structures of the two networks are radically different, sharing a few edges and thus explaining the low precision and recall scores observed in \textcolor{black}{all results presented in this work}. 
Interestingly, some children nodes of the ground truth BN are identified as parents in the solution found by NSGA-II (e.g., variable $V_2$) and vice versa (e.g., variable $V_5$). 
We stress the fact that, in real world scenarios, the structure of the ground truth is completely unknown; therefore, the log-likelihood represents the only viable estimation of the quality of the putative solutions. 
Thus, the best individuals found by NSGA-II would be accepted as optimal solutions to the problem.

\textcolor{black}{W}e repeated our experiments on a real world BN that is commonly used in literature as a benchmark, i.e., the  causal signaling networks presented by Sachs \emph{et al}. \cite{sachs2005causal}, which is characterized by $11$ variables corresponding to the proteins PKC, PKA, Raf, Mek, Erk, Akt, Jnk, P38, PIP2, PIP3 and Plcg. 
In this test, we performed $30$ inferences of the BN. 
During each run we exploited $50$ samples, to test the most difficult scenario for the algorithms.
The results of this test are shown in Figure \ref{fig:sachs} and confirm the validity of our previous findings: NSGA-II was able to identify a Pareto front that dominates both HC and the ground truth  from the point of view of log-likelihood, retaining the same complexity in terms of number of arcs. 
Interestingly, in this test, the average performance of HC for both  AIC and BIC \textcolor{black}{regularizator} is the same.

\textcolor{black}{Finally, we want to briefly compare and discuss the computational complexity of the MOO algorithm employed in this work, with respect to the classic HC. 
The most computationally expensive part of a BN inference process is the evaluation of the likelihood, whose complexity is proportional to the number of samples $m$ in the dataset. 
The HC algorithm performs an iterative stochastic movement in the space of feasible BNs, in which new arcs can be added or existing arcs can be removed, following the direction that maximizes the likelihood. 
This process is repeated for a fixed amount of iterations,\ -- which was set to $10\;000$ in our experiments -- and a   likelihood evaluation is performed during each iteration. 
In the case of NSGA-II, the overall number of likelihood evaluations is equal to the number of fitness evaluations  which, in turn, is equal to $Q \times IT_{\texttt{max}} = 12 \, 800$.
Hence, the overall computational effort due to likelihood calculations of the two methods is roughly the same; however, NSGA-II carries the additional complexity of the fast non-dominated sorting, whose complexity is $\mathcal{O}(\Sigma Q^2)$, and all the calculations due to the evolutionary process.
}

\section*{Conclusion}

\label{sec:conclusion}
In this paper we \textcolor{black}{extensively investigated and analysed the performance of} NSGA-II, a  multi-objective optimization algorithm, \textcolor{black}{when solving the problem of learning the structure of Bayesian Networks}. 
\textcolor{black}{The ostensible advantage of using} NSGA-II to simultaneously optimize the likelihood and the number of arcs\textcolor{black}{, is that it does not require} any heuristic to regularize the likelihood score. 
To the best of our knowledge, such \textcolor{black}{problem formulation of identifying a} 
set of optimal solutions \textcolor{black}{for the BN structure, which} cannot be further improved without an effect in one of the fitness values, hence providing a comprehensive characterization and explicit trade-off between likelihood and model complexity\textcolor{black}{, was never thoroughly examined}. 

\textcolor{black}{Such formulation of the optimization problem, together with the tests we performed,} provide evidence for the superiority of this optimization strategy and, in fact, NSGA-II is shown to be more effective in optimizing the two objective functions, when compared to HC. 
We also tested SPEA2 as multi-objective optimization algorithm, to highlight that our results are consistent regardless the actual  meta-heuristics that is employed.

\textcolor{black}{The main finding of this paper concerns the solutions found by multi-objective optimization algorithms. As a matter of fact, it} is interesting to note that, although characterized by dominated fitness values, the solutions obtained by regularizators such as BIC or AIC are typically closer to the structure of the (unknown) ground truth network that was used to generate the dataset employed as input for the optimization process, \textcolor{black}{with respect to the results obtained with a multi-objective optimization algorithm}. 
This state of affairs is explainable by the fact that the search strategy to solve the structural learning task of BNs comprising HC (or similar approaches) paired with a regularized likelihood score is, on the one hand, conservative as it results in a low number of selected arcs leading to sparse solutions; on the other hand, this approach guarantees the selection of only the handful of arcs that are more strongly supported by (or, state otherwise, that are contributing more to) the likelihood score. 
However, \textcolor{black}{we stress the fact that} in real world scenarios, the objective functions represent the only available measures of the quality of the solutions and they are used to discriminate the optimal solution of the problem. 
\textcolor{black}{Indeed, the results presented in this work, in which NSGA-II achieved better results than HC from the objective functions point of view, despite the optimized structure of the BN can be drastically different from the ground truth, underline} the complexity of the inference problem especially when BNs are used for their causal interpretation. 

Moreover, NSGA-II is also shown to outperform AIC and BIC in the case of highly connected BNs, and it is more robust to noise in the dataset, possibly due to the mutation operator which allows a better exploration of the space of feasible solutions, preventing the premature convergence to local minima. 
Our results represent a step toward a better understanding of the limitations of likelihood-based approaches to be further investigated in future research. 

\bibliographystyle{abbrv}
\bibliography{regulbn}

\end{document}